# Deep learning enabled multi-wavelength spatial coherence microscope for the classification of malaria-infected stages with limited labelled data size

*Neeru Singla[1], and Vishal Srivastava[1]*

[1]*Department of Electrical and Instrumentation Engineering, Thapar Institute of Engineering and Technology, Patiala, Punjab 147004, India*

**Corresponding author:** *vishalsrivastava17@gmail.com*

Malaria is a life-threatening mosquito-borne blood disease hence early detection is very crucial for health. The conventional method for the detection is a microscopic examination of Giemsa-stained blood smears, which needs a highly trained skilled technician. Automated classifications of different stages of malaria still a challenging task especially having poor sensitivity in detecting the early trophozoite and late trophozoite or schizont stage with limited labelled datasize. The study aims to develop a fast, robust and fully automated system for the classification of different stages of malaria with limited data size by using the pre-trained convolutional neural networks (CNNs) as classifier and multi-wavelength to increase the sample size. We also compare our customized CNN with other well-known CNNs and shows that our network have a comparable performance with less computational time. We believe that our proposed method can be applied to other limited labelled biological datasets.

## 1. Introduction

Malaria is a mosquito-borne disease that affected many people in the worldwide generation [1]. According to 2015 statistics, 214 million malaria-infected cases were reported causing an approximate death toll of 438,000 worldwide [1], [2]. It is the most common syndrome which requires the proper treatment at a very early stage. Early and efficient diagnosis of malaria makes it preventable as well as curable. Different stage of malaria-infected red blood cells (RBCs) are early and late trophozoite. In general, in early trophozoite stage no change in size and two or more chromatin dot present on it while in late trophozoite sturdy, vacuolated and dark pigment occur. To assess the impact of promising antimalarial drug and characterizing the distinctive life cycle phases of malaria parasites on host cell invasion process, parasite departure and schizont development [3]. Such a framework will encourage the present endeavors went for recognizing new antimalarial medications, and antibodies focusing on the distinctive erythrocytic phases of Plasmodium falciparum (*P. falciparum*) parasites. Due to the low cost and renowned method, microscopic testing is the gold standard technique for the diagnosis of malaria [4]–[7]. Over the last decades, significant progress has been made in the field of quantitative phase microscopy to overcome the limitation of traditional phase microscopy [8]–[10]. Several quantitative phase imaging (QPI) methods such as optical diffraction tomography, digital holography quantitative phase microscopy/spectroscopy, refractive index tomography are used for the quantification of unstained malaria-infected RBCs [9], [13] – [16]. Although, all these techniques are very much successful in biomedical applications but they also have certain disadvantages. To capture the dynamic behaviour of the biological cells such as membrane fluctuations, dry mass cell density etc. an off-axis interferometry or holography is preferred since it gives complete information in a single shot whose response solely depends upon recording device speed [17] – [19] . However, coherent noise and parasitic fringes formation due to highly spatially and temporal coherent light source (laser light) reduces the quality of images which results in phase measurement inaccuracy [19]. To overcome this problem broad band light sources such as white light, and LED is generally used for which have a high spatial phase sensitivity due to low temporal coherence length [21] – [23]. Due to low temporal coherence length, the interference will only occur when the path difference between the reference and sample arm is within the coherence length which limits the utilization of whole camera field of view [23]. The low fringe density also makes it difficult a single-shot phase recovery, therefore, a multi-phase-shifting algorithm is utilized for phase extraction at full detector resolution [24] which limits its application for the study of dynamic behaviour of the cells. The other disadvantage with a broadband light source-based system is that it required chromatic aberration corrected optical components [25], [26]. However, these issues can be resolved with the use of high temporal and low spatially coherent light source. The major advantage of this kind of light source (high temporal and low spatially coherent) that it has a high fringe density over the whole camera field of view and it doesn't require any dispersion compensation mechanism. Therefore, this source is more suitable for the dynamic biological sample which has a strong dispersion or inhomogeneous spectral response.  The cells have different absorption, emission and scattering properties with respect to each wavelength. Using multi-wavelength imaging will helpful in extracting the different information embedded in it. As refractive index is a function of wavelength and is directly correlated with structural and biomechanical characteristics of the sample [27].

In the past few years, several efforts have been made to use machine learning for automatic detection of malaria infection from microscopic images of stained blood cells to avoid human interpretation error [4], [6], [7]. Most of the machine learning methods such as K-NN, Naive Bayes, ANN, and SVM etc. uses shape measurement, color features and statistical features for the classification of malaria parasite-infected RBCs [4], [28], [29] but their achieved accuracy varies from 84–95% in detecting parasites from stained blood cells [4], [5], [15]. As discussed, though there are attempts to automate the malaria diagnosis based on morphological features but there sensitivity and specificity due to overlapping color intensities makes it difficult to classify early and late trophozoite as shown in Fig. 1.

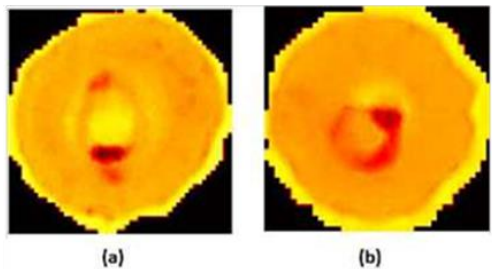

**Figure 1** Bright field images of (a) Early trophozoite and (b) Late trophozoite RBCs, respectively.

In the case of images, the major source of information is the spatial local correlation among the neighboring pixels. In that case CNN a class of deep learning (DL) will be a good option for the classification [30], [31].  In medical imaging analysis and interpretation CNN's, a branch of DL, have an excellent record over other approaches [30], [32].  Thus this will become an obvious choice for the analysis of images. Different arrangement of the filters will lead to different architecture of CNN. Biomedical domain generally suffers from the limited amount of data and these data also contain high variability which will cause "overfitting". Due to overfitting, the features can't generalize well on data and training DL model from scratch requires extensive memory and large computational power, which limits its application in the biomedical domain. To overcome this problem "transfer learning" and "fine-tuning" would be a good solution.

In this paper, we demonstrated how a customized CNN with a multi-wavelength spatial coherence microscope (SCM) can be used for automatic identification of different stages of malaria with limited dataset. Moreover, multi-wavelength imaging helps to increase the sample size, which is necessary for data-hungry DL for automatic classification of the cells. This approach is utilized for the classification of healthy and infected (early and late trophozoite) RBCs as well as early and late trophozoite RBCs. We analyze the performance of our developed system with different pre-trained CNN's (AlexNet, VGGNet, GoogleNet, ResNet and Inception models, and customized CNN). The present system with customized CNN has a comparable performance with less computational time in detecting different stages of malaria. Our approach, multi-wavelength SCM with deep learning has not been used in the medical imaging literature so far.

## 2. Experimental Details

For analyzing the malaria-infected RBCs, the SCM system is used. The schematic diagram of the system is presented in Fig. 2, which includes three lasers (632 nm, 532 nm and 460 nm), two beam splitters ($BS_1$, $BS_2$), two microscopic objective lens ($MO_1$, $MO_2$), multi-mode fiber bundle (MMFB), CCD camera and neutral density filter (ND). At a time, one light source is used only. The red laser (632 nm) is incident onto the beam splitter ($BS_1$), which splits the beam into two parts, one part is going to $MO_1$ and another part is going towards $MO_2$. Further, $MO_2$ is coupled with 50/50 fiber-based beam splitter. Similarly, the blue and green laser also illuminated the sample one at a time. ND filter is placed in the path to equalize the density of all the three beams. The three beams illuminates the rotating diffuser at an angle of $+40^0$, $0^0$, $-40^0$ respectively, as shown in the Fig. 2 to make speckle patterns statistically independent which will reduce the speckle contrast effectively [33]. The spot size on the diffuser plate is ~6 cm. The MMFB with a core diameter 10 mm and contains 100 fibers (0.1 mm core diameter each fiber) collected the output light of diffuser.

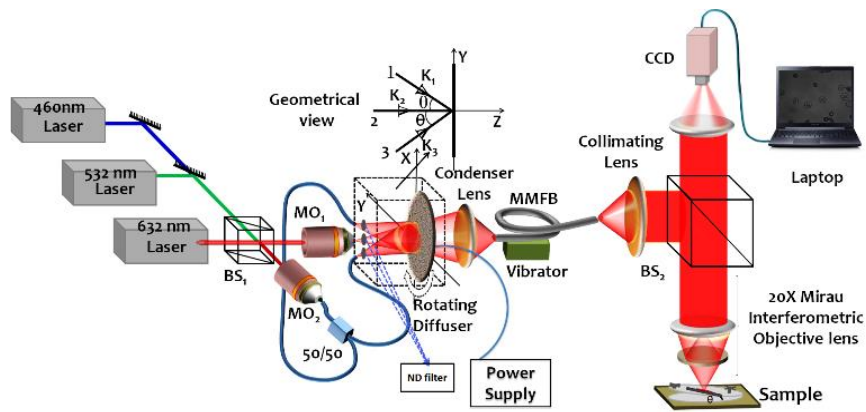

**Figure 2** Schematic diagram of SCM system.

The collimating lens of focal length 17.5 mm collected the scattered output light of the MMFB and fed to the Nikon microscope (Nikon Eclipse 50i). The emerging collimated light was incident on 20X Mirau interferometer (Edmund Optics, NA 0.4, WD 4.7 mm), the transmitted light incident onto the sample placed a slightly off-axis mirror, while the reflected beam is focused to an aluminized spot on the front surface of the microscope objective. The two beams recombine at $BS_2$ and the interference pattern is recorded by the CCD camera [SamBa EZ-140, Sensovation AG, 1392×1040 pixels with pixel size 4.65 × 4.65 $\mu m^2$, fps = 30 and full well capacity=14500 e-]. The field of view of the SCM image at CCD is 400 μm x 400 μm.

## 3. Materials And Methodology

### 3.1 Principle

In coherence theory of optical fields, the temporal coherence function is determined by Wiener-Khintchin theorem while spatial coherence function is determined by Van-Cittert Zernike theorem [34]. According to Wiener-Khintchin theorem, for a broad light source spectrum source frequency spectrum is the Fourier transform of the temporal coherence function [35]. Similarly, according to Van-Cittert Zernike theorem, source spatial distribution is the Fourier transform of the spatial coherence function [34]. For spatially longitudinal incoherent light source, the coherence length is found by longitudinal spatial coherence (LSC) length instead of temporal coherence as in the case of low coherence interferometry [34], [36]. The LSC length of the spatially extended light source is defined by

$$\gamma_{long}(z_1, z_2) = \int_{-\infty}^{\infty} S(k_z) \exp(ik_z \delta z) dk_z \qquad (1)$$

where, $\gamma_{long}(z_1, z_2)$ = LSC of the light source,

$S(k_z)$ = angular spectrum of the light field

$k_z$ = longitudinal spatial frequency

$\delta z = (z_1 - z_2)$, difference between two-spatial points $O_1(z_1)$ and $O_2(Z_2)$ in the observation plane as shown in Fig. 3 (a).

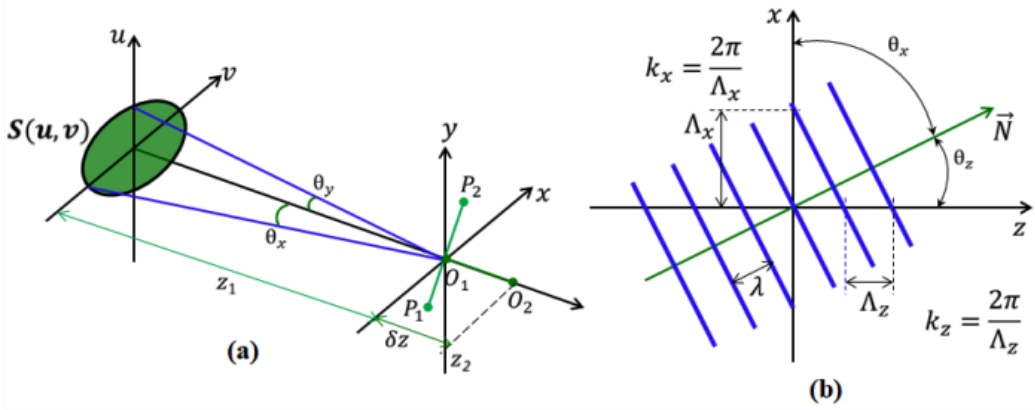

**Figure 3** (a) Spatial coherence to the extended light source, (b) spatial periods and circular frequencies of a plane wave.

$$K_z = \frac{2\pi}{\Lambda_z} = \frac{2\pi}{\lambda} \cos\theta_z \qquad (2)$$

where $\Lambda_z$ is a spatial period along Z-direction as shown in Fig. 3(b), $\lambda$ is the wavelength of the light source and $\theta_z$ is the angle between the z-axis and direction of propagating field,

The longitudinal coherence length is defined by

$$L_c = \frac{2\pi}{\Delta k_z} \qquad (3)$$

where $\Delta k_z$ is the longitudinal spatial frequency range.
The longitudinal coherence frequency, which depends upon the temporal frequency as well as angular frequency is given by

$$L_c = \left[\frac{2\sin^2\left(\frac{\theta_z}{2}\right)}{\lambda_0} + \frac{\Delta\lambda}{\lambda_0^2}\cos^2\frac{\theta_z}{2}\right]^{-1} \quad (4)$$

where, $\lambda_0$ = central wavelength of the light source,

$\Delta\lambda$ = temporal spectrum width of the source

$\theta_z$ = half of the angular spectrum width.

For monochromatic light source $(\Delta\lambda \ll \lambda_0)$ expression (4) becomes

$$L_c = \frac{\lambda_0}{2\sin^2\left(\frac{\theta_z}{2}\right)}$$

(4)

hence, the LSC of the light source can be purely determined by the angular spectrum width of the light source [37].

The lateral resolution is computed by

$$L_S \cong \frac{0.61 \times \lambda_0}{NA} \quad (5)$$

Thus the longitudinal coherence length is dominated by LSC rather than temporal coherence of the light source. To calculate the axial resolution of the experimental set-up we place a mirror as a sample and move the sample arm in the positive and negative axial direction till the fringe visibility losses. We extract the envelope of the recorded fringe contrast and the full-width half maxima is the axial resolution. It comes out to be 4.5 μm, 4 μm and 3.8 μm for the red, green and blue wavelength in the air, respectively for 20X (0.4 NA) Mirau interferometer. To measure the lateral resolution USAF test target is placed in the sample arm and record the image with 20X (0.4 NA), the 6th element of the 7th group has been resolved. The lateral resolution is 1.9 μm, 1.1 μm and 1 μm for the red, green and blue wavelength in the air, respectively. The calculated phase sensitivity of the system is ~10m radian.

### *3.2    Sample Preparation*

All the blood samples were taken from the Ganga pathology, Gorakhpur, India with informed written consent. The samples were stored in Ethylene diamine tetraacetic acid (EDTA) anticoagulant tubes. To maintain the cellular microenvironment the RBCs were initially diluted in phosphate buffer saline (pH~7.4) solution and centrifuged and then give some time for the cell settlement.

### *3.3    RBC Patch Extraction*

Entropy of each image of size n×n is calculated with the help of equation no (6).

$$Entropy = -\sum_{i=1}^{L} P_i \log P_i \quad (6)$$

where, L is the maximum phase value, $P_i$ is the probability of each phase value in the phase image, which can be obtained from equation no (7).

$$P_i = \frac{f(i,j)}{n^2} \tag{7}$$

where $f(i,j)$ is the i[th] histogram count and $n^2$ is the number of pixels in the patch of an image.

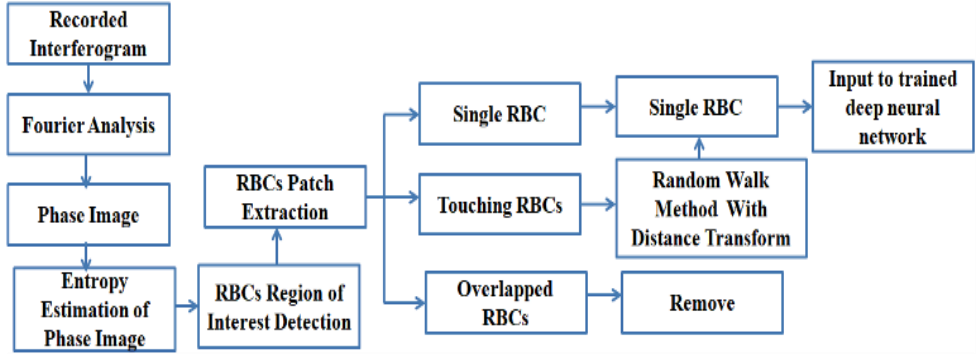

**Figure 4** Flowchart of RBC patch extraction algorithm.

The entropy threshold is set to be 1.0 radian obtained by the number of trails and the higher entropy region is extracted i.e. RBC region. Now remove all those artifacts whose area is less than 40 x 40 pixels$^2$. To RBC patch is extracted with the help of the minimum coordination of pixel and maximum pixel coordination from the boundary of the pixel. There is a number of patches which consist of touching and overlapping RBCs. Overlapping RBCs are excluded from the study and the RBCs which are touching are separated with the help of random walk method [38] in association with the distance transform [39]. More details can be found on [40] and the whole process is explained in Fig. 4.

*3.4 Data Augmentation*

DL is datasets hungry, which is generally difficult in the medical field. Our dataset is relatively small (1000 images). However, in our case, we develop the QPI system using multi-wavelength red, green and blue which will three times the sample size (3 RGB phase images). Therefore, with the help of our system, we augmented data [42], [43]. To further increase the datasets the images were rotate ($45^0$, and $135^0$) but make sure that the numbers of samples for each class of RBCs must be same thereby balancing out the dataset.

*3.5 Methodology*

The whole description of the methodology used in this manuscript to identify healthy vs infected and early vs late trophozoite malaria-infected stages is shown in Fig. 5.

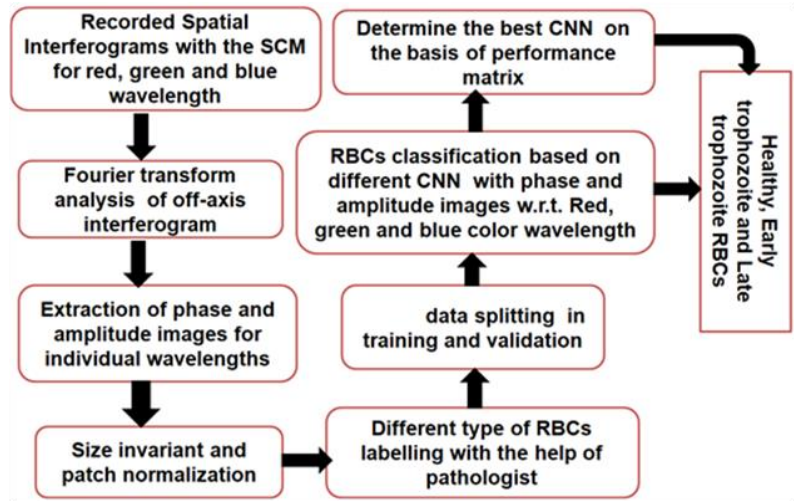

**Figure 5** Overall flowchart of our proposed training and learning methodology for the malaria infected RBCs CNN classification model.

Spatial interferograms were recorded using the different wavelengths (red, green and blue) of three light sources (one light source used at a time) from the region of interest (ROI) to generate complex field (both amplitude and phase ($\Delta\phi(x, y)$)). Once, the slightly off-axis spatial interferograms were captured for red wavelength. The captured spatial interferogram was initially Fourier transformed then positive or negative first order spectrum was filtered out by spatially filtered and inverse Fourier transformation was executed to obtain amplitude and wrapped phase. Unwrap phase map is obtained by Goldstein's method [41]. The same process has been repeated for the green and blue wavelength.

*3.6    Models Description*

In this study, we exploit the different layers architecture of CNN for the classification purpose such as AlexNet, VGG16, ResNet50, GoogLeNet, and custom-made network.

**AlexNet:** operates on input images of size 227x227x3. It has five convolution layers, three pooling layers, three fully connected layers and also includes dropouts [44] .

**VGG-16:** similar architecture as AlexNet but consist more convolution layers. It operates on the input image of size 224x224x3. It is composed of 13 convolution layers, 5 pooling layers, 3 fully connected layers and along with activation rectified linear unit (ReLU) and dropouts [45].

**ResNet50:** is a 50 layer Residual Network. This network introduces is residual learning, which tries to learn residual instead of learning some features. Residual are nothing but subtraction of learned feature from the input of that layer [46].

**GoogLeNet:** builds on input size 299x299x3. It includes four convolution layers along with batch normalization (BN) and activation ReLU, four pooling layers and nine inception layers which have convolution blocks followed by batch normalization and activation ReLU and also includes pooling. The inception module is an intrinsic component of GoogLeNet [47].

**Customized CNN network:** A number of CNN architectures are designed with a different number of convolutional layers, and each layer has different size (3x3, 5x5) and a different number of filters. Each convolutional layer is alternation with ReLu and pooling layers and a different number of filters. To minimize overfitting a dropout layer is added. The proposed CNN topology is implemented in Python (on Tensor Flow Platform). The proposed CNN model and the optimization parameters are specified in Table 1.

Table 1 Customized CNN Architecture

| Layer | Type | Input | Kernel | Filters | Stride | Pad | Activations | Output |
|---|---|---|---|---|---|---|---|---|
| Data | Input | 120x120x3 | N/A | N/A | N/A | N/A | N/A | 120x120x3 |
| Conv1 | Convolution | 120x120x3 | 3x3 | 64 | 1 | 1 | ReLU | 120x120x64 |
| Conv2 | Convolution | 120x120x64 | 3x3 | 96 | 2 | 2 | ReLU | 61x61x96 |
| Pool2 | Max Pooling | 61x61x96 | 2x2 | - | 2 | 0 | - | 30x30x96 |
| Conv3[a] | Convolution | 30x30x96 | 2x2 | 128 | 2 | 1 | ReLU | 16x16x128 |
| Conv4 | Convolution | 16x16x128 | 3x3 | 256 | 1 | 1 | ReLU | 16x16x256 |
| Pool4 | Max Pooling | 16x16x256 | 2x2 | - | 2 | 0 | - | 8x8x256 |
| Conv5 | Convolution | 8x8x256 | 3x3 | 256 | 2 | 1 | ReLU | 5x5x256 |
| fc6[b] | Fully Connected | 5x5x256 | 9x9 | 1000 | 1 | 0 | Tanh | 1000x1 |
| fc7[c] | Fully Connected | 1000x1 | 1x1 | 1 | 1 | 0 | Sigmoid | 1 |

a. Trained using dropout rate of 20%
b. Trained using dropout rate of 50%
c. Trained using L2-regularization

The proposed model consists of five convolution layers and two fully connected layers. To introduce non-linearity and to make fast convergence learning ReLU layer is added to all the convolution and fully connected layers. All the negative activation values become zero by ReLU layer. To reduce the spatial dimension max pooling layer is added to the second and fourth convolution layer after ReLU layer. Gaussian distribution with standard deviation 0.01 is used to initiate all the network weights for all the layers. To improve the performance of the network, dropout layer with probability (0.2) is added after third convolution layer and probability (0.5) is added after first fully-connected layer. L2 regularization technique with a value of $\lambda = 10^{-3}$ is used for weight decay to avoid network from overfitting of the training data [48]. The learning rate is 0.0001 which after every 4 epochs decays by using inverse decay policy during training. Stochastic gradient descent (SGD) is used to optimize the loss function with a batch size of 32. To avoid any impact on learning all the dataset were randomly shuffled. To avoid loss function to trap into local minima and to move global minima, the momentum factor is set to 0.9. All the models were pre-trained on natural image dataset ImageNet and fine-tuned on our training dataset. The training and validation sets were randomly split.

## 4. Results and Discussion

In this study, an automated TL model was used to discriminate malaria-infected stages (early and late trophozoite) of RBCs from the healthy RBCs. To evaluate the performance of CNN used for the classification of healthy vs infected RBCs and early vs late trophozoite RBCs several experiments have been performed. Figure 6 shows the phase images for healthy, early trophozoite and late trophozoite for red, green and blue wavelengths, respectively. In total 36 subjects (8 healthy, 15 early trophozoite and 13 late trophozoite) were used for analysis. The sample was split based on subjects into three non-overlapping subsets: to train the dataset 5 healthy, 10 early trophozoite and 7 late trophozoite subject was used, for validation 2 healthy, 3 early trophozoite and 4 late trophozoite subject was used and to test the dataset 1 healthy, 2 early trophozoite and 2 late trophozoite subject was used. For training and validation, 7 healthy, 13 early trophozoite and 11 late trophozoite subject datasets were used while for testing, 1 healthy, 2 early trophozoite and 2 late trophozoite subject datasets were used. The testing dataset is saved in a different folder and doesn't take a part in the training or validation of the network.

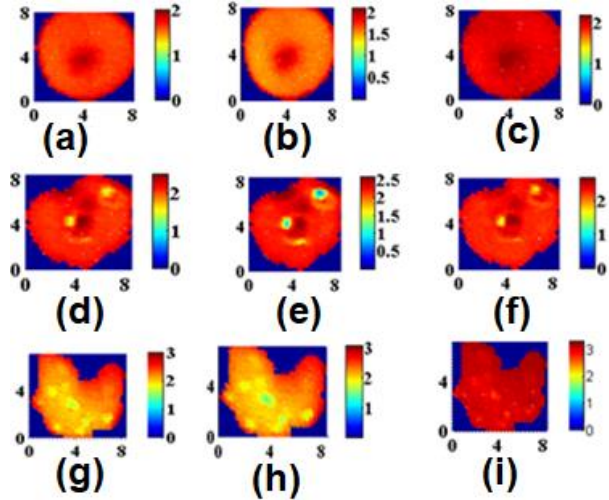

**Figure 6** (a), (b), (c) are the phase (radian) images of healthy RBCs for red, green and blue wavelength, respectively, (d), (e), (f) are the phase (radian) images of early trophozoite RBCs for red, green and blue wavelength, respectively, and (g), (h), (i) are the phase (radian) images of late trophozoite RBCs for red, green and blue wavelength, respectively.

Further, to evaluate each network performance by feeding only phase information w. r. t. three wavelengths as a three-input channel. For each sample of the three wavelengths (red, green and blue), we have recorded 10 fields of view (FoV) at the different location of the same slide. The ground truth for the images was determined by a trained person who is expert in parasite identification. There are number of RBCs that even the trained pathologist was not able to find out the difference between early and late trophozoite from the bright field images. These phase images were removed from the training, but still used in testing. The total RBCs phase images extracted from training and validation dataset were – 3045 out of which 632 healthy RBCs, 602 early trophozoite RBCs and 611 late trophozoite RBCs, per subject for each wavelength were selected and analyzed. For dataset balancing, we chose 600 RBCs of each stage and the total number of RGB phase images RBCs images are 5400. To increase the robustness of the network data augmentation technique is also applied ($45^0$ and $135^0$ rotation). Hence, the total number of sample size for training and validation is 16200. The testing is performed on 350 phase images extracted from healthy, early and late trophozoite RBCs. We cropped the small patch of size 60 x 60 pixels$^2$ phase images, which contains the single RBC. All RBC patch images were normalized to the same size 60 x 60 pixels$^2$. A different model has different receptive fields, therefore, we upsampled the size of cropped images according to network requirement. All three wavelengths phase images were used at the input channel of the network.

Therefore, the final size of the image at the network input is m x n x 3, where m and n are the number of input pixels to the different network. The input data was enhanced with the help of data augmentation technique. Different types of data augmentation techniques are used to improve the robustness of RBCs classification CNN model.

**Table 2** Performance metric for the Identification of Healthy and Infected RBCs

| Models | Sensitivity | Specificity | Accuracy | MCC |
|---|---|---|---|---|
| AlexNet | 0.971 | 0.963 | 0.971 | 0.941 |
| VGG16 | 0.981 | 0.966 | 0.973 | 0.947 |
| **ResNet50** | **0.983** | **0.969** | **0.976** | **0.952** |
| GoogLeNet (Inception-V3) | 0.981 | 0.967 | 0.974 | 0.949 |
| **Customized CNN** | **0.979** | **0.965** | **0.972** | **0.945** |

**Table 3** Performance metric for the Identification of Early and Late Trophozoite RBCs

| Models | Sensitivity | Specificity | Accuracy | MCC |
|---|---|---|---|---|
| AlexNet | 0.901 | 0.903 | 0.901 | 0.803 |
| VGG16 | 0.901 | 0.904 | 0.903 | 0.806 |
| ResNet50 | 0.906 | 0.903 | 0.904 | 0.809 |
| **GoogLeNet** | **0.907** | **0.917** | **0.912** | **0.824** |
| **Customized CNN** | **0.903** | **0.907** | **0.905** | **0.810** |

In our experiment, all the transfer learning models and custom model is trained using SGD with a momentum factor of 0.9 and the learning rate for the pre-trained network is 0.0001. All the models were trained for 15 epochs and the minibatch size for AlexNet and VGG16 is 64 whereas, for ResNet50, GoogLeNet and customized network the minibatch size is 32 due to memory constraints. All the models were trained and tested on a Windows 10 system with Intel (R) Xeon (R) CPU E5-1620 0 3.60-GHz processor, 1 TB HDD, 16 GB RAM, a CUDA-enabled Nvidia GeForce GTX 670 12GB GPU, and CUDA 9.0 dependencies for GPU acceleration.

**Table 4** Computational Test Time of all the Models

| Network | Test time per image (ms) |
|---|---|
| AlexNet | 38 |
| VGG16 | 70 |
| ResNet50 | 208 |
| GoogleNet(Inception-V3) | 348 |
| Customized CNN | 17 |

A comparison study was conducted on healthy vs infected RBCs (early and late trophozoite RBCs) and early vs late trophozoite RBCs using different deep CNN's training model. Table 2 shows the performance metrics of all the models for the classification of healthy and infected (early and late trophozoite) RBCs, while table 3 shows performance metrics of all the models for the classification of early and late trophozoite RBCs for testing dataset. The results present that in table 2 and 3, shows that Customized CNN model has a comparable performance with the other transfer learning model with less number of layers. The customized network has a lower run-time which overall decreases the computational time and increases the throughput with respect to time as mentioned in table 4.

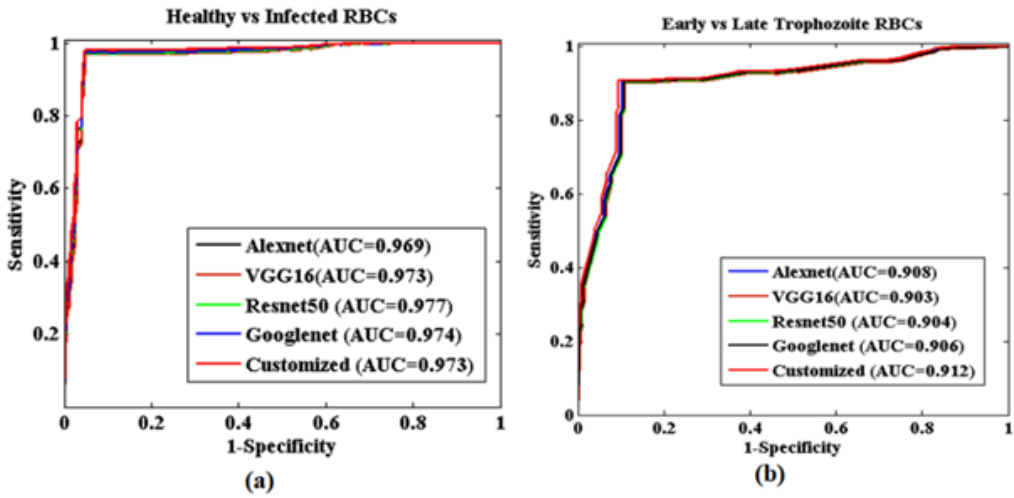

**Figure 7** AROC for testing dataset of all the models (a) Healthy vs malaria infected RBCs and (b) Early vs late trophozoite malaria infected stages of RBCs.

Sensitivity and Specificity in the previous studies [15], [29] using QPI and machine learning is high in the case of healthy and malaria-infected RBCs (95% - 98%) however, in the case of malaria-infected stages (early and late trophozoite) achieved poor sensitivity i.e. 45.0% - 66.8% and 99.8% specificity [9]. In our study, we achieved 97.7% accuracy, 98.3% sensitivity and 96.9% specificity, respectively for healthy and malaria infected RBCs whereas in the case of malaria-infected stages between early and late trophozoite stages 91.2% accuracy, 90.3% sensitivity and 90.7% specificity, respectively. The area under the receiver operating characteristic curve (AROC) for all the models is shown in Fig. 7. Figure 7(a) shows the AROC for all the models for healthy vs infected RBCs and Fig. 7(b) shows the AROC for all the models for early vs late trophozoite RBCs. The customized network performance with less number of layers is comparable to the other deeper networks. In future, a more accurate segmentation algorithm will be used to incorporate overlapping RBCs and the present system will be modified to get the spectral response of the cells which will strengthen the performance for the classifying the sickle cells, diabetics and cancer cells etc.

## 5. Conclusions

In summary, we have designed an automated, robust, high-throughput multi-wavelength SCM system based on low spatial coherence for the classification of different stages of malaria stages using DL. The purpose multi-wavelength SCM with CNN promises of serving as an effective diagnostic system, where there is a limited labelled data sample. We achieved 97.7% accuracy, 98.3% sensitivity and 96.9% specificity, for the healthy vs malaria infected RBCs, whereas in the case of malaria-infected stages between early vs late trophozoite stages 91.2% accuracy, 90.3% sensitivity and 90.7% specificity, respectively. The results show the robustness of the system with limited labelled data size. We strongly assume that this study can be dominant to significantly improve the screening accurate results for other health-related issues.

**Disclosure**

The author declares that there are no conflicts of interest.

**Acknowledgements**

The authors are thankful to Prof. D. S. Mehta, Indian Institute of Technology Delhi for providing his lab facilities to conduct experiment.